\documentclass[preprint,12pt]{elsarticle}

\usepackage{amssymb}
\usepackage{amsmath}
\usepackage{psfrag}
\usepackage{epsfig}
\usepackage{graphics}
\usepackage{graphicx}
\usepackage{color}

\usepackage{subfigure}
\usepackage{multirow}
\usepackage{threeparttable}
\usepackage{booktabs}
\usepackage{bbm}
\usepackage[T1]{fontenc}
\usepackage{algorithm}
\usepackage{algpseudocode}
\usepackage{amsmath}
\usepackage{multirow}
\usepackage{psfrag,epsfig,subfigure}
\usepackage{float}
\usepackage{booktabs}

\begin{document}

\begin{frontmatter}


\title{ImplantFormer: Vision Transformer based Implant Position Regression Using Dental CBCT Data}

\author[label1,label2,label3]{Xinquan Yang\corref{equalcontribution}}
\ead{xinquanyang99@gmail.com}
\author[label4]{Xuguang Li\corref{equalcontribution}}
\ead{lixuguang@szu.edu.cn}
\author[label1,label2,label3]{Xuechen Li}
\ead{timlee@szu.edu.cn}
\author[label5]{Peixi Wu}
\ead{WPX564@163.com}
\author[label1,label2,label3]{Linlin~Shen\corref{mycorrespondingauthor}}
\ead{llshen@szu.edu.cn}
\author[label4]{Yongqiang Deng\corref{mycorrespondingauthor}}
\ead{qiangyongdeng@sina.com}
\cortext[mycorrespondingauthor]{Corresponding author}
\cortext[equalcontribution]{Equal contribution}

\address[label1]{College of Computer Science and Software Engineering, Shenzhen University, Shenzhen, China}
\address[label2]{AI Research Center for Medical Image Analysis and Diagnosis, Shenzhen University, Shenzhen, China}
\address[label3]{National Engineering Laboratory for Big Data System Computing Technology, Shenzhen University, China}
\address[label4]{Department of Stomatology, Shenzhen University General Hospital, Shenzhen, China}
\address[label5]{the School of Dentistry, Shenzhen University, Shenzhen, China}

\begin{abstract}
Implant prosthesis is the most appropriate treatment for dentition defect or dentition loss, which usually involves a surgical guide design process to decide the implant position. However, such design heavily relies on the subjective experiences of dentists. In this paper, a transformer-based Implant Position Regression Network, ImplantFormer, is proposed to automatically predict the implant position based on the oral CBCT data. We creatively propose to predict the implant position using the 2D axial view of the tooth crown area and fit a centerline of the implant to obtain the actual implant position at the tooth root. Convolutional stem and decoder are designed to coarsely extract image features before the operation of patch embedding and integrate multi-level feature maps for robust prediction, respectively. As both long-range relationship and local features are involved, our approach can better represent global information and achieves better location performance. Extensive experiments on a dental implant dataset through five-fold cross-validation demonstrated that the proposed ImplantFormer achieves superior performance than existing methods.
\end{abstract}




\end{frontmatter}


\section{Introduction}\label{sec1}
Periodontal disease is the world's 11th most prevalent oral condition, which causes tooth loss in adults, especially the aged ~\cite{elani2018trends}~\cite{nazir2020global}. Implant prosthesis is so far the most appropriate treatment of dentition defect/dentition loss, and using surgical guide in implant prosthesis leads to higher accuracy and efficiency ~\cite{varga2020guidance}~\cite{vinci2020accuracy}~\cite{gargallo2021intra}. Cone-beam computed tomography (CBCT) and oral scanning are common data used for surgical guide design, among which the CBCT data is used to estimate the implant position, while the oral scanning data is employed to analyze the surface of the teeth. However, such a process takes the dentist a long time to analyze the patient's jaw, teeth and soft tissue using surgical guide design software. Artificial Intelligence (AI) based implant position estimation could significantly speed up such process~\cite{amato2013artificial}.

Recently, deep learning has achieved great success in many tasks of dentistry~\cite{schwendicke2021artificial}~\cite{muller2021barriers}~\cite{kim2023automatic}~\cite{chen2021hierarchical}. For dental implant planning, recent researchers mainly focus on implant depth estimation. Kurt et al.~\cite{kurt2021deep} utilised multiple pre-trained convolutional networks to segment the teeth and jaws to locate the missing tooth and generate a virtual tooth mask according to the neighbouring teeth' location and tilt. The implant depth is determined by measuring the width of the alveolar bone, the distance of the mandibular canal, maxillary sinus and jaw bone edge of the virtual mask using the panoramic radiographic image. Widiasri et al. introduced Dental-YOLO~\cite{widiasri2022dental} to detect the alveolar bone and mandibular canal on the sagittal view of CBCT to determine the height and width of the alveolar bone. However, we argue that only measuring the depth cannot well determine the implant position. In contrast, the 2D axial view of CBCT is more appropriate for implant position estimation since the precise implant position can be obtained by stacking multiple 2D axial views. Nevertheless, in the clinical, the implant is inserted into the alveolar bone to act as the tooth root, in which the attached soft tissues around the tooth root will lead to blurry CT images, which makes a big challenge to implant position estimation.

In this paper, we propose to train the prediction network using the 2D axial view of the tooth crown, which is exposed in the air and can be clearly captured by CT imaging. To obtain the implant position label at the tooth crown, we first fit the centerline of the implant using tooth root annotations and then extend the centerline to the crown area (see Fig.~\ref{fig_prosedure}(b)). By this means, the implant position at the tooth crown area can be obtained and used as labels to train the prediction network. During inference, the outputs of the network will be transformed back to the tooth root area, as the predicted positions of the implant (see Fig.~\ref{fig_prosedure}(d)). Moreover, the fitted centerline combines the prediction results of a series of 2D slices of CBCT, which compensates the loss of 3D context information without introducing heavy computation costs. Inspired by the dentist who determines the implant position with reference to the neighboring teeth~\cite{kurt2021deep}~\cite{liu2021transfer}, we employ Vision Transformer (ViT)~\cite{dosovitskiy2020image} as the backbone for our Implant Position Regression Network (ImplantFormer). By dividing the input image into equal-sized patches and applying multi-head self-attention (MHSA) on image patches, all pixels can establish a relationship with others, which is extremely important for implant position decisions with reference to the texture of neighboring teeth. To further improve the performance of ImplantFormer, we design a convolutional stem to coarsely extract the image feature before patch embedding and then extract multi-level features from the backbone and introduce a convolutional decoder for further feature transformation and fusion.

The main contributions of this paper are summarized as follows.
\begin{itemize}
\item We creatively propose to predict the implant position using the 2D axial view of the tooth crown area, which has better image quality than the tooth root and predicts a precise implant position.
\item The centerline of the implant is fitted using a series of prediction results on the 2D axial view, which introduces the 3D context information without requiring heavy computation costs.
\item A transformer-based Implant Position Regression Network (ImplantFormer) is proposed to consider both local context and global information for more robust prediction.
\item The experimental results show that the proposed ImplantFormer achieves superior performance than the mainstream detectors.
\end{itemize}


\section{Related work}\label{sec2}
\subsection{Deep Learning in Dentistry}\label{subsec2}
Deep learning technology is widely used in many tasks of dentistry, e.g., dental caries detection, 2D and 3D tooth segmentation, and dental implants classification. For dental caries detection, Schwendicke et al.~\cite{schwendicke2020deep} and Casalegno et al.~\cite{casalegno2019caries} proposed deep convolutional neural networks (CNNs) for the detection and diagnosis of dental caries on Near-Infrared-Light Transillumination (NILT) and TI images, respectively. For tooth segmentation, Kondo et al.~\cite{kondo2004tooth} proposed an automated method for tooth segmentation from 3D digitized image captured by a laser scanner, which avoids the complexity of directly processing 3-D mesh data. Xu et al.~\cite{xu20183d} presented a label-free mesh simplification method particularly tailored for preserving teeth boundary information, which is generic and robust in the complex appearance of human teeth. Lian et al.~\cite{lian2020deep} integrated a series of graphconstrained learning modules to hierarchically extract multiscale contextual features for automatically labeling on raw dental surface. Cui et al.~\cite{cui2021tsegnet} proposed a two-stage segmentation method for 3D dental point cloud data. The first stage uses a distance-aware tooth centroid voting scheme to ensure the accurate localization of tooth and the second stage designs a confidence-aware cascade segmentation module to segment each individual tooth and resolve the ambiguous cases. Qiu et al.~\cite{qiu2022darch} decomposed the 3D teeth segmentation task as the tooth centroid detection and tooth instance segmentation, which provides a novel dental arch estimation method and introduces an arch-aware point sampling (APS) module based on the estimated dental arch for tooth centroid detection. For the task of implant fixture system classification, Sukegawa et al.~\cite{sukegawa2020deep} evaluated the performance of differnet CNN models for implant classification. Kim et al.~\cite{kim2020transfer} proposed an optimal pretrained network architecture for identifying different types of implants.

\subsection{Deep Learning in Object Detection}\label{subsec2}
The object detectors can be divided into two categories, i.e., anchor-based and anchor-free. The anchor-based detector sets the pre-defined anchor box before training and the anchor-free detector directly regresses the bounding box of the object. Furthermore, the anchor-based detector can be grouped into one-stage and two-stage methods. YOLO~\cite{redmon2016you} is a classical one-stage detector, which directly predict the bounding box and category of objects based on the feature maps. A series of improved versions of YOLO~\cite{bochkovskiy2020yolov4}~\cite{li2022yolov6}~\cite{wang2023yolov7}~\cite{ge2021yolox} have been proposed to improve the performance. Faster R-CNN~\cite{ren2015faster} is a classical two-stage detector that consists of a region proposal network (RPN) and a prediction network (R-CNN~\cite{girshick2014rich}). Similarly, a series of detection algorithms~\cite{cai2018cascade}~\cite{sun2021sparse}~\cite{he2017mask} have been proposed to improve the performance of the two-stage detector. Different from the anchor-based detector that heavily relies on the predefined anchor box, the anchor-free detector regresses the object using the heatmap. CornerNet~\cite{law2018cornernet} simplified the prediction of the object bounding box as the regression of the top-left corner and the bottom-right corner. CenterNet~\cite{duan2019centernet} further simplified CornerNet by regressing the center of object. Recently, transformer-based anchor-free detector achieves great success in object detection. DETR~\cite{carion2020end} employs ResNet as the backbone and introduces a transformer-based encoder-decoder architecture for the object detection task. Deformable DETR~\cite{zhu2020deformable} extends DETR with sparse deformable attention that reduces the training time significantly.

\subsection{Deep Learning in Implant Position Estimation}\label{subsec3}
The computer-aided diagnosis (CAD) systems were applied to dental implant planning earlier~\cite{amato2013artificial}. Pol{\'a}{\v{s}}kov{\'a} et al.~\cite{polavskova2013clinical} presented a web-based tool which utilized patient history and clinical data input into a program and preset threshold levels for various parameters to formulate a decision on whether or not implants may be placed. Sadighpour et al.~\cite{sadighpour2014application} developed an ANN model which utilized a number of input factors to formulate a decision regarding the type of prosthesis (fixed or removable) and the specific design of the prosthesis for rehabilitation of the edentulous maxilla. Szejka et al.~\cite{szejka2011reasoning} developed an interactive reasoning system which requires the dentist to select the region of interest within a 3D bone model based on computed tomography (CT) images, to help the selection of the optimum implant length and design. However, these CAD systems need manual hyperparameter adjustment.

Recently, researchers proposed different approaches to determine the implant position using the panoramic radiographic images and 2D slices of CBCT. Kurt et al.~\cite{kurt2021deep} utilised multiple pre-trained convolutional networks to segment the teeth and jaws to locate the missing tooth and determine the implant loaction according to the neighbouring teeth' location and tilt. Widiasri et al. introduced Dental-YOLO~\cite{widiasri2022dental} to detect the alveolar bone and mandibular canal based on the sagittal view of CBCT to determine the height and width of the alveolar bone, which determines the implant position indirectly.

\begin{figure}
\centering
\includegraphics[width=1.0\linewidth]{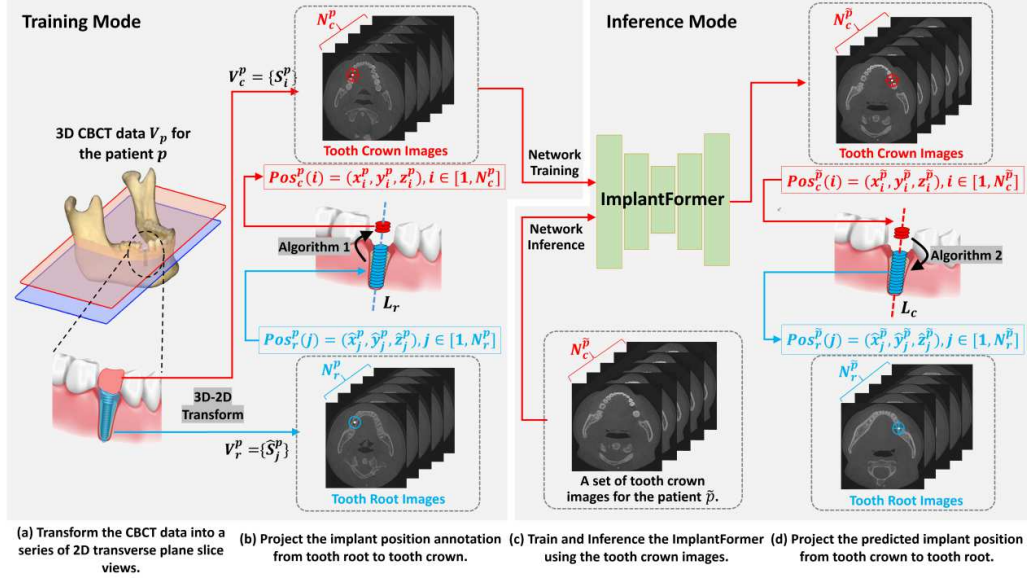}
\caption{The whole procedure of the proposed implant position prediction method, which consists of training and inference mode.}\label{fig_prosedure}
\end{figure}

\section{Method}
The whole procedure of the proposed implant position prediction method is shown in Fig.~\ref{fig_prosedure}, which consists of four stages: 1) Firstly, we transform the 3D CBCT data $V_p$ of patient $p$ into a series of 2D transverse plane slice views, $V^p_c=\{S^p\}$ and $V^p_r=\{{\hat S}^p\}$. As the 3D implant regression problem is now modeled as a series of 2D regression of $Pos^p$, the prediction network can now be simplified from 3D to 2D and greatly alleviate the insufficiency of training data; 2) Use proposed Algorithm 1 to project the ground truth annotation $Pos^p_r$ from tooth root to tooth crown $Pos^p_c$; 3) Train ImplantFormer using crown images $S^p_i$ with projected labels and predict the implant position of the new patient $\tilde{p}$ at tooth crown; 4) in the end, Algorithm 2 is proposed to transform the predicted results $Pos^{\tilde{p}}_c$ back from tooth crown to tooth root $Pos^{\tilde{p}}_r$.

\begin{figure}
\centering
\includegraphics[width=1.0\linewidth]{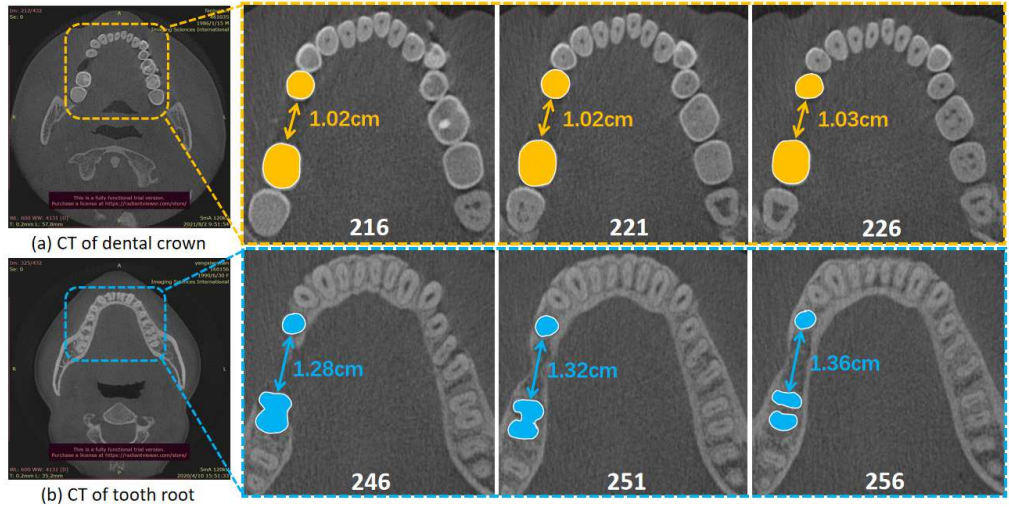}
\caption{Visual comparison of the 2D slice of tooth crown (first row) and tooth root (second row). The white number represents the serial number of slice. Areas covered in yellow and blue represent the total area of neighboring teeth at tooth crown and tooth root, respectively; and arrows show the distance between the neighboring teeth.} \label{fig_compare_root_crown}
\end{figure}

\subsection{CBCT Data Pre-processing}
As shown in Fig.~\ref{fig_prosedure}(a), the 3D CBCT data of the patient $p$, $V^p$ consists of two parts, i.e. tooth crown data $V^p_c$ and tooth root data $V^p_r$, $V^p=V^p_c\cup V^p_r$. Here $V^p_c=\{S^p_i\},i \in [1,N^p_c]$ and $V^p_r=\{\hat S^p_j\},j \in [1,N^p_r]$, $S^p_i$ and $\hat S^p_j$ are the 2D slices of tooth crown and tooth root, respectively. $N^p_c$ and $N^p_r$ are the total number of 2D slices of tooth crown and tooth root, respectively.

In the clinical, the implant is inserted into the alveolar bone to act as the tooth root. However, the 2D slices captured from the tooth root usually look blurry and the shape of tooth root varies much more sharply across slices, which makes a big challenge to the training of the prediction network. In contrast, the texture in 2D slices corresponding to tooth crown areas are much richer and more stable. As shown in Fig.~\ref{fig_compare_root_crown}, with an interval of 10 slices, the distance between two neighboring teeth at tooth crown only increases by 0.01 centimeter while which is 0.08 at the tooth root. Meanwhile, the textures of tooth area (including the implant position and its neighboring teeth) in the crown images are richer and more stable than that of tooth root, which is beneficial for the network regression. More detailed comparison of tooth crown and root images are given in the experimental section.

Hence, in this work, we propose to train the prediction network using crown images - $V^p_c$. However, the ground truth position of implant, defined as $Pos^p_r(j)=(\hat x^p_j,\hat y^p_j,\hat z^p_j),j \in [1,N^p_r]$, is annotated in $V^p_r$. Here $\hat x^p_j,\hat y^p_j$ are the coordinates of implant position in $\hat S^p_j$ and $\hat z^p_j$ is the slice index of $\hat S^p_j$ in $V^p_r$. To obtain the implant position annotation $Pos^p_c$ in $V^p_c$, a space transform algorithm $T_{{Pos^p_r}\rightarrow{Pos^p_c}}$ is proposed. As shown in Fig.~\ref{fig_prosedure}(b), firstly, we use $Pos^p_r$ to fit the center line of implant, defined as $L_r$; then $L_r$ is extended to the crown area and the intersections of $L_r$ with $V^p_c$, $Pos^p_c$, can be obtained. The detailed algorithm is given in Algorithm~\ref{alg1}. The input and output of the algorithm are $Pos^p_r$ and $Pos^p_c$, respectively. Given the slice index of $S^p_i$ as $z^p_i$, we define a space transform $T_{{Pos^p_r}\rightarrow{Pos^p_c}}$ to calculate the $x^p_i$ and $y^p_i$. Here $k_1$, $k_2$, $b_1$ and $b_2$ can be calculated by minimizing the residual sum of square $Q$.

\begin{figure}
\centering
\includegraphics[width=1.0\linewidth]{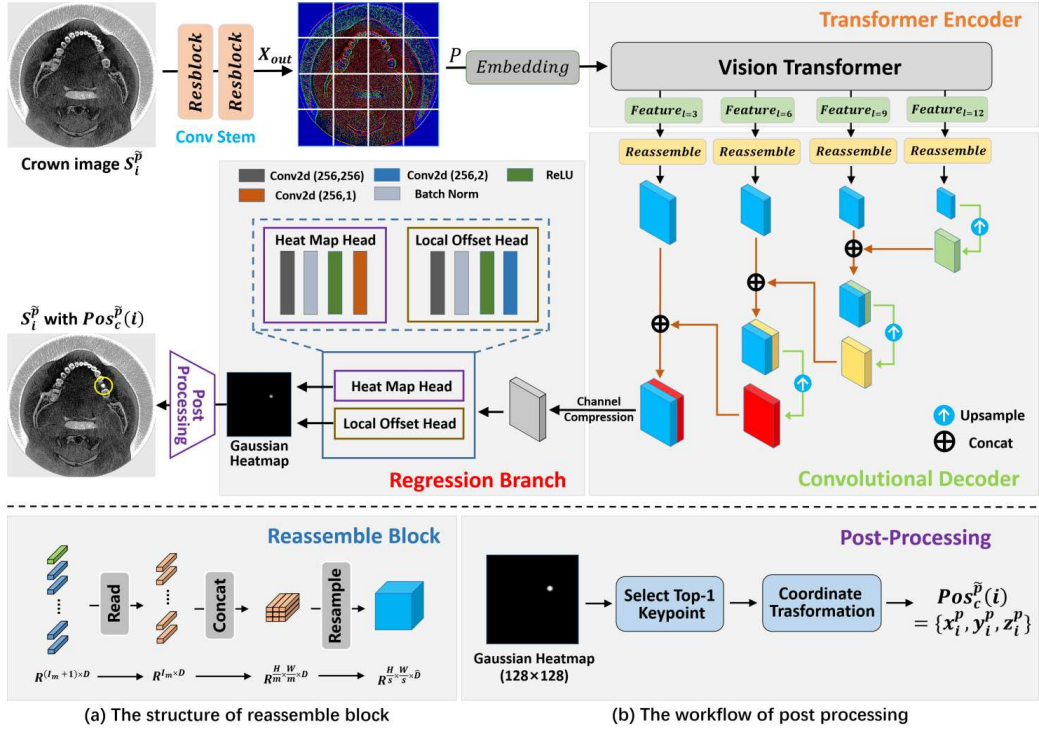}
\caption{The network structure of the proposed ImplantFormer.
}
\label{fig_network}
\end{figure}

\subsection{Transformer Based Implant Position Regression Network (ImplantFormer)}
The input and output of the ImplantFormer is the 2D slice at tooth crown $S_i \in R^{H\times W\times 3}, (H=W=512)$ and the corresponding implant position $Pos_c(i)$, respectively. Similar to the existing keypoint regression network~\cite{zhou2019objects}~\cite{law2018cornernet}, ImplantFormer is based on the Gaussian heatmap and the center of the implant position $(x_i, y_i)$ at 2D slice is set as the regression target. The network structure of ImplantFormer is given in Fig.~\ref{fig_network}, which mainly consists of four components: convolutional stem, transformer encoder, convolutional decoder and regression branch. Given a tooth crown image $S^{\tilde{p}}_i \in R^{H\times W\times 3}$ of the patient $\tilde{p}$ as the network input, we firstly introduce two Resblocks~\cite{he2016deep} as the convolutional stem for coarsely extracting the image feature, and then the general patch embedding is generated using the output of convolutional stem. Subsequently, we employ ViT-Base-ResNet-50 model as the transformer encoder and extract multi-level features. In convolutional decoder, we introduce reassemble blocks~\cite{ranftl2021vision} to transform the multi-level features to multi-resolution feature maps and then integrate them by upsampling and concatenation. Finally, focal loss~\cite{lin2017focal} and L1 loss are employed to supervise the heatmap head and the local offset head, respectively. The output of ImplantFormer is the implant positions at the tooth crown, which is extracted from the predicted heatmap by the post processing operation.

\subsubsection{Convolutional Stem} Adding a convolutional stem before transformer encoder is demonstrated to be helpful in extracting finer feature representation~\cite{zhu2020deformable}~\cite{meng2021conditional}~\cite{carion2020end}. To this end, we design a convolutional stem for the ImplantFormer, which consists of two Resblocks. Unlike previous works~\cite{wang2022anchor}~\cite{dai2021up}~\cite{li2022dn} that use the last stage feature of CNN backbone as the input of transformer encoder, we keep the channel and size of the output tensor $X_{out} \in \mathbb{R}^{H\times W\times 3}$ the same as the input image, so that the original ViT can be directly used as backbone.

\subsubsection{Transformer Encoder} ViT strikes a remarkable success in computer vision, due to the property of modeling long-ranged dependencies. In our work, the regression of implant position heavily relies on the texture of the neighboring teeth, which is similar to the way dentist used to design the surgical guide. Consequently, the prediction network should possess the capacity of establishing relationship between long-ranged pixels, i.e. between the implant position and the neighboring teeth. ViT divides the input image into equal-sized patches $P \in R^{m\times m\times 3}$, and patch embedding is applied by flattening the patch into vectors using a linear projection. Then, multi-head self-attention (MHSA) is applied to establish relationship between different image patches. By this means, each pixel can perceive the adjacent pixels even with long distance. This characteristic is essential for our work that uses the texture of the neighboring teeth for implant position prediction.

Hence, in this work, we employ ViT-Base-ResNet-50 model as backbone for our ImplantFormer. The non-overlapping square patches are generated using the output of the convolutional stem. The patch size  is set as 16 for all experiments. Meanwhile, to further improve the performance of ImplantFormer, we extract multi-level features from layers $l=\{3,6,9,12\}$. The produced multi-level features with 256 dimensions are used to generate image-like feature representations for the convolutional decoder.

\subsubsection{Convolutional Decoder} Unlike transformer-based encoder-decoder structures, we introduce the convolutional decoder for further feature transformation and fusion. As shown in Fig.~\ref{fig_network}, the reassemble block~\cite{ranftl2021vision} firstly transform the multi-levels ViT features to multi-resolutions feature maps and then the image-like feature maps are integrated by upsampling and concatenation. The final concatenated feature map is fed into the $3\times3$ convolution for feature smooth and channel compression.

Specifically, we employ a simple three-stage reassemble block to recover image-like representations from the output tokens of arbitrary layers of the transformer encoder (shown as Fig.~\ref{fig_network}(a)), the reassemble operation is defined as follows:
\begin{equation}
Reassemble_{s}^{\hat{D}}(t)=(Resample_{s} \circ Concat \circ Read)(t),
\end{equation}
where $s$ denotes the output size ratio of the recovered representation with respect to the input image, $\hat D$ denotes the output feature dimension, and $t$ refers to the readout token from transformer encoder.
\begin{equation}
Read=R^{{I_m+1}\times D}\rightarrow R^{I_m\times D},
\end{equation}
\begin{equation}
Concat=R^{{I_m}\times D}\rightarrow R^{{\frac{H}{W}}\times {\frac{H}{W}}\times D},
\end{equation}
\begin{equation}
Resample_s=R^{{\frac{H}{W}}\times {\frac{H}{W}}\times D}\rightarrow R^{{\frac{H}{s}}\times {\frac{H}{s}}\times \hat D}.
\end{equation}

As shown in the above equations, reassemble block consists of three steps: Read, Concat and Resample. Read operation firstly map the $I_m+1$ tokens to a set of $I_m$ tokens, where $I_m=\frac{HW}{m^2}$ and $D$ is the feature dimension of each token. After the Read block, the resulting $I_m$ tokens are reshaped into an image-like representation by placing each token according to the position of the initial patch in the image. Then, a spatial concatenation operation is applied to produce a feature map of size $\frac{H}{m} \times \frac{W}{m}$ with channels. Finally, a spatial resampling layer is proposed to scale the representation to size $\frac{H}{s} \times \frac{W}{s}$ with $\hat D$ features per pixel.

After the reassemble operation, a simple feature fusion method is used to integrate the multi-resolution feature maps. We upsample the feature map by a factor of two and concatenate it with the neighboring feature map. For the final concatenated feature map, we use $3\times3$ convolution for feature smoothing and reducing the channel from 512 to 256.

\subsubsection{Regression Branch} The outputs of ImplantFormer is an implant position heatmap $F\in[0,1]^{\frac{W}{g} \times \frac{H}{g}}$, where $g$ is the down sampling factor of the prediction and set as 4. The heatmap $F$ is expected to be equal to 1 at the center of implant position, and equal to 0 otherwise. Following the standard practice in CenterNet~\cite{zhou2019objects}, the probability of pixel $(x,y)$ being the target center point is modeled as a 2D Gaussian kernel:
\begin{equation}
\mathcal{L}_k=\frac{-1}{N}\sum_{xy}\left\{
\begin{array}{ccl}
(1-\hat F_{xy})^\alpha\log(\hat F_{xy}) & \text{if $F_{xy}=1$}, \\
(1-\hat F_{xy})^\beta\log(\hat F_{xy})^\alpha \log(1-\hat F_{xy}) & \text{otherwise},
\end{array}\right.
\end{equation}
where $\alpha$ and $\beta$ are hyper-parameters of the focal loss, $\hat F$ is the predicted heatmap and $N$ is the number of keypoints in image. We use $\alpha=2$ and $\beta=4$ in all our experiments. To further refine the prediction location, the local offset head is used for each target center point, which is optimized by L1 loss. The loss of the local offset $\mathcal{L}_{off}$ is weighted by a constant $\lambda_{off}=0.55$. The overall training loss is:
\begin{equation}
\mathcal{L}_{total}=\mathcal{L}_k+\lambda_{off}\mathcal{L}_{off}.
\end{equation}

\begin{algorithm}
  \caption{Space transform from $Pos^p_r$ to $Pos^p_c$.}\label{alg1}
  \label{alg:Framwork}
  \begin{algorithmic}
    \Require
    The groundtruth annotation of implant position in $V^p_r$: $Pos^p_r(j)=(\hat x^p_j,\hat y^p_j,\hat z^p_j),j\in [1,N^p_r]$.

    \Ensure
    The annotation of implant position in $V^p_c$: $Pos^p_c(i)=(x^p_i,y^p_i,z^p_i),i\in [1,N^p_c]$.
  \end{algorithmic}
  \begin{enumerate}
    \item Define a space transform $T_{Pos^p_r \to Pos^p_c}$ based on $Pos^p_r$
    \begin{equation*}
    T_{Pos^p_r \to Pos^p_c}(\hat z^p_j) = \left\{
    \begin{array}{rl}
    &\hat x^p_j=k_1\hat z^p_j+b_1 \\
    &\hat y^p_j=k_2\hat z^p_j+b_2
    \end{array} \right.
    \end{equation*}
    where $k$ and $b$ are the coefficient and bias of $L_r$, respectively.

    \item Perform the residual sum of squares $Q$ on $T_{Pos^p_r \to Pos^p_c}$:
    \begin{equation*}
    Q_1=\sum^{N^p_r}_{j=1}(\hat x^p_j-k_1\hat z^p_j-b_1)^2, Q_2=\sum^{N^p_r}_{j=1}(\hat y^p_j-k_2\hat z^p_j-b_2)^2
    \end{equation*}
    \item Employ the least square method to minimize $Q$, calculate the derivative of $Q$ with respect to $k$ and $b$, respectively, and set them to 0.
    \begin{equation*}
    \frac{\partial Q_1}{\partial k_1}=\sum^{N^p_r}_{j=1}2(\hat x^p_j-k_1\hat z^p_j-b_1)\times(-\hat z^p_j)=0
    \end{equation*}
    \begin{equation*}
    \frac{\partial Q_1}{\partial b_1}=\sum^{N^p_r}_{j=1}2(\hat x^p_j-k_1\hat z^p_j-b_1)\times(-1)=0
    \end{equation*}
    \begin{equation*}
    \frac{\partial Q_1}{\partial k_2}=\sum^{N^p_r}_{j=1}2(\hat y^p_j-k_2\hat z^p_j-b_2)\times(-\hat z^p_j)=0
    \end{equation*}
    \begin{equation*}
    \frac{\partial Q_1}{\partial b_2}=\sum^{N^p_r}_{j=1}2(\hat y^p_j-k_2\hat z^p_j-b_2)\times(-1)=0
    \end{equation*}
    \item Use $Pos^p_r$ to calculate $k$ and $b$.
    \begin{equation*}
    k_1=\frac{N^p_r \sum_{j=1}^{N^p_r} \hat x_j\hat z_j-\sum_{j=1}^{N^p_r} \hat x_j \times \sum_{j=1}^{N^p_r} \hat z_j}{N^p_r \sum_{j=1}^{N^p_r} z_j^2-\sum_{j=1}^{N^p_r} z_j \times \sum_{j=1}^{N^p_r} z_j},
    b_1=\frac{\sum_{j=1}^{N^p_r} \hat x_j \times k_1\sum_{j=1}^{N^p_r} \hat z_j}{N^p_r}
    \end{equation*}
    \begin{equation*}
    k_2=\frac{N^p_r \sum_{j=1}^{N^p_r} \hat y_j\hat z_j-\sum_{j=1}^{N^p_r} \hat y_j \times \sum_{j=1}^{N^p_r} \hat z_j}{N^p_r \sum_{j=1}^{N^p_r} z_j^2-\sum_{j=1}^{N^p_r} z_j \times \sum_{j=1}^{N^p_r} z_j},
    b_2=\frac{\sum_{j=1}^{N^p_r} \hat y_j \times k_2\sum_{i=1}^{N^p_r} \hat z_j}{N^p_r}
    \end{equation*}
    \item Substitute $k$, $b$ and $z^p_j$ into $T_{Pos^p_r \to Pos^p_c}$ to obtain $Pos^p_c$.
    \begin{equation*}
    Pos^p_c(i)=(k_1z^p_i+b_1,k_2z^p_i+b_2,z^p_i),i\in [1,N^p_c]
    \end{equation*}
  \end{enumerate}
\end{algorithm}

\begin{algorithm}
  \caption{Space transform from $Pos^{\tilde{p}}_c$ to $Pos^{\tilde{p}}_r$.}\label{alg2}
  \label{alg:Framwork}
  \begin{algorithmic}
    \Require
    The outputs of ImplantFormer for patient $\tilde{p}$: $Pos^{\tilde{p}}_c(i)=(x^{\tilde{p}}_i,y^{\tilde{p}}_i,z^{\tilde{p}}_i),i\in [1,N^{\tilde{p}}_c]$.

    \Ensure
    The predicted implant position at tooth root: $Pos^{\tilde{p}}_r(j)=(\hat x^{\tilde{p}}_j,\hat y^{\tilde{p}}_j,\hat z^{\tilde{p}}_j),j\in [1,N^{\tilde{p}}_r]$.
  \end{algorithmic}
  \begin{enumerate}
    \item Define a space transform $T_{Pos^{\tilde{p}}_c \to Pos^{\tilde{p}}_r}$ based on $Pos^{\tilde{p}}_c$
    \begin{equation*}
    T_{Pos^{\tilde{p}}_c \to Pos^{\tilde{p}}_r}(\hat z^{\tilde{p}}_i) = \left\{
    \begin{array}{rl}
    &x^{\tilde{p}}_i=\hat k_1z^{\tilde{p}}_i+\hat b_1 \\
    &y^{\tilde{p}}_i=\hat k_2z^{\tilde{p}}_i+\hat b_2
    \end{array} \right.
    \end{equation*}
    where $\hat k$ and $\hat b$ are the coefficient and bias of $L_c$, respectively.

    \item Perform the residual sum of squares $\hat Q$ on $T_{Pos^{\tilde{p}}_c \to Pos^{\tilde{p}}_r}$:
    \begin{equation*}
    \hat Q_1=\sum^{N^{\tilde{p}}_r}_{i=1}(\hat x^{\tilde{p}}_i-k_1\hat z^{\tilde{p}}_i-b_1)^2, \hat Q_2=\sum^{N^{\tilde{p}}_r}_{i=1}(\hat y^{\tilde{p}}_i-k_2\hat z^{\tilde{p}}_i-b_2)^2
    \end{equation*}
    \item Calculate the derivative with respect to $k$ and $b$ on $\hat Q$, and set them to 0.
    \item Use $Pos^{\tilde{p}}_c$ to calculate $\hat k$ and $\hat b$.
    \item Substitute $\hat k$, $\hat b$ and $\hat z^{\tilde{p}}_j$ into $T_{Pos^{\tilde{p}}_c \to Pos^{\tilde{p}}_r}$ to obtain $Pos^{\tilde{p}}_r$.
    \begin{equation*}
    Pos^{\tilde{p}}_c(j)=(\hat k_1\hat z^{\tilde{p}}_j+\hat b_1,\hat k_2\hat z^{\tilde{p}}_j+\hat b_2,\hat z^{\tilde{p}}_j),j\in [1,N^{\tilde{p}}_c]
    \end{equation*}
  \end{enumerate}
\end{algorithm}

\subsubsection{Post-processing} The output size of the heatmap is smaller in scale than that of the input image, due to the down sampling operation. We introduce a post-processing to extract the implant position from the heatmap and recover the scale of output. As shown in Fig.~\ref{fig_network}(b), the post-processing includes two steps: top-1 keypoint selection and coordinate transformation. We firstly select the prediction result with the highest confidence. Then, the Gaussian heatmap with the selected implant positions are directly transformed from the resolution of $128\times128$ to $512\times512$. The coordinate of implant position can be obtained by extracting the brightest point in the Gaussian heatmap.

\subsection{Project Implant Position from Tooth Crown to Tooth Root}
After the post processing, a set of implant position at tooth crown $Pos^{\tilde{p}}_c$ are obtained. To obtain the real location of implant at tooth root $Pos^{\tilde{p}}_r$, we propose Algorithm~\ref{alg2} to transform the prediction results back from tooth crown to tooth root. Algorithm~\ref{alg2} has the same workflow as Algorithm~\ref{alg1}, except that the input and output are reversed. As shown in Fig.~\ref{fig_prosedure}(d), firstly, we use $Pos^{\tilde{p}}_c$ to fit the space line $L_c$; then $L_c$ is extended to the root area and the intersections of $L_c$ with $V^{\tilde{p}}_r$, $Pos^{\tilde{p}}_r$, can be obtained. The detailed algorithm is given in Algorithm~\ref{alg2}. Moreover, the fitted centerline combines the prediction results of a series of 2D slices of CBCT, which compensates the loss of 3D context information without introducing heavy computation costs.

\section{Experiments and Results}
\subsection{Dataset Details}
\begin{figure}
\centering
\includegraphics[width=1.0\linewidth]{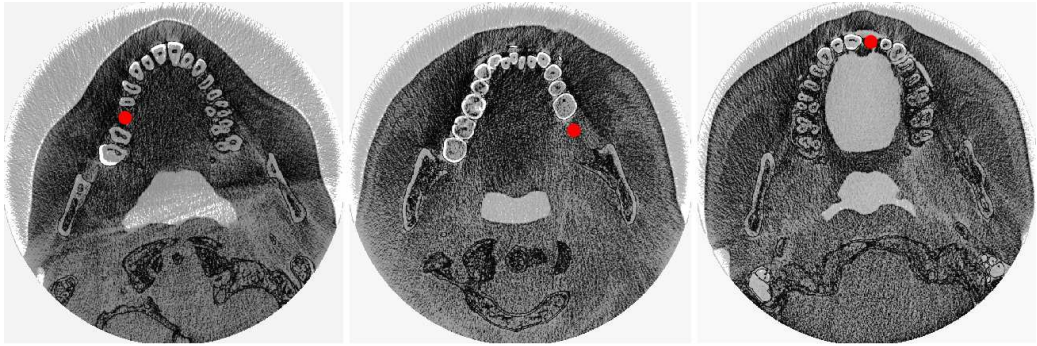}
\caption{Some sample images in our dataset. The red points denote the implant position annotation.} \label{fig_dataset}
\end{figure}

We evaluate our method on a dental implant dataset collected from the Shenzhen University General Hospital (SUGH), which contains 3045 2D slices of tooth crown and the implant position was annotated by three experienced dentists. These CBCT data were captured using the KaVo 3D eXami machine, manufactured by Imagine Sciences International LLC. Dentists firstly designed the virtual implant based on the CBCT data using the surgical guide design software. Then the implant position can be determined as the center of the virtual implant. Some sample images of the dataset are shown in Fig.~\ref{fig_dataset}.


\begin{figure}
\centering
\includegraphics[width=1.0\linewidth]{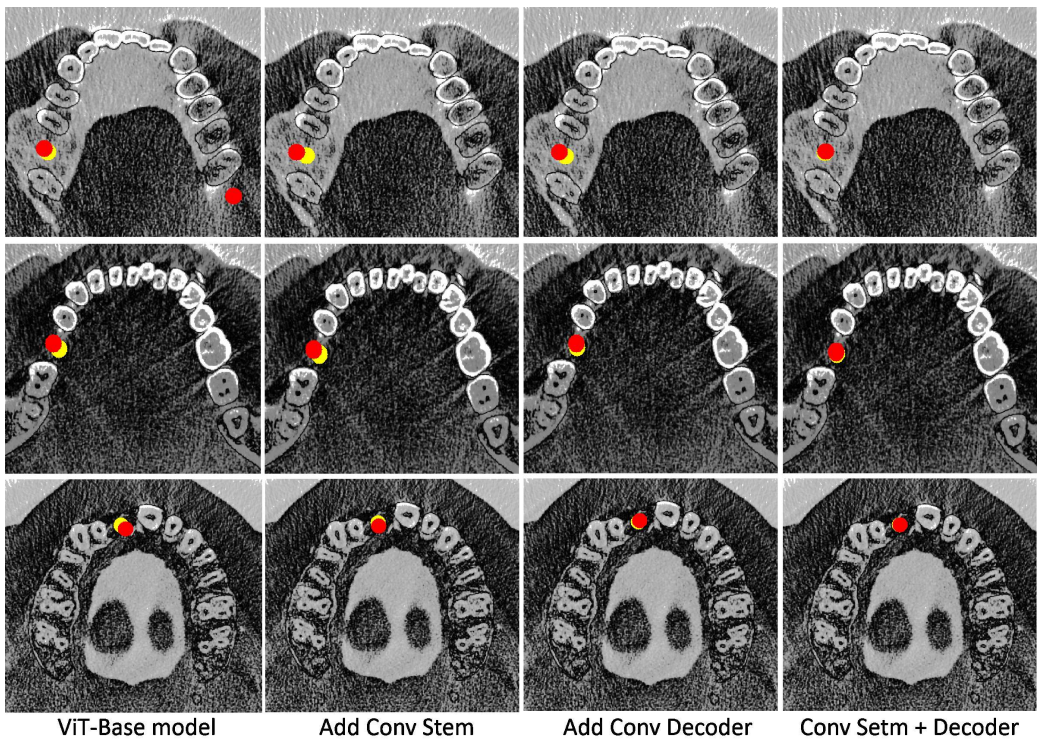}
\caption{Visual comparison of detection results of the proposed component. The red and yellow circles represent the predicted implant position and ground truth position, respectively.} \label{fig_vis_compare}
\end{figure}

\subsection{Implementation Details}
For our experiments, we use a batch size of 6, Adam optimizer and a learning rate of 0.0005 for network training. The network is trained for 140 epochs and the learning rate is divided by 10 at 60th and 100th epochs, respectively. Three data augmentation methods, i.e. random crop, random scale and random flip are employed in the network training. The original images are with size $776\times776$ and cropped to $512\times512$ at the network training stage. In the inference stage, images are directly resized to $512\times512$. All the models are trained and tested on the platform of NVIDIA GeForce RTX TITAN. Five-fold cross-validation is employed for all experiments.

\subsection{Evaluation Criteria}
The diameter of the implant is 3.5$\sim$5mm, and clinically the mean error between the predicted and ideal implant position is required to be less than 1mm, i.e., around 25\% of the size of implant. Therefore, AP75 is used as the evaluation criteria. As the average radius of implants is around 20 pixels, a bounding-box with size $21\times21$ centered at the keypoint is generated. The calculation of AP is defined as follows:
\begin{equation}
Precition=\frac{TP}{TP+FP}
\end{equation}
\begin{equation}
Recall=\frac{TP}{TP+FN}
\end{equation}
\begin{equation}
AP=\int^1_0P(r)dr
\end{equation}

Here TP, FP and FN are the number of correct, false and missed predictions, respectively. P(r) is the PR Curve where the recall and precision act as abscissa and ordinate, respectively.

\subsection{Performance Analysis}
\subsubsection{Tooth Crown vs. Tooth Root} To justify the effectiveness of tooth crown images for implant position estimation, we train the ImplantFormer using tooth crown and root images, respectively, and compare their performances in Table~\ref{table2}. We can observe from the table that when setting the IOU value as 0.5 and 0.75, the prediction of ImplantFormer using crown images achieves 34.3\% and 13.7\% AP, respectively, which is 24.9\% and 13.1\% higher than using tooth root images. The experimental results demonstrate the efficiency of using tooth crown images for implant position prediction.

\begin{table}
\caption{The detection performance of ImplantFormer trained using tooth crown and tooth root images.}\label{table2}
\centering
\begin{tabular}{ccc}
\toprule
CBCT slice  & $AP_{50}\%$    & $AP_{75}\%$   \\ \hline
Tooth crown & \textbf{34.3}$\pm$\textbf{2.2684} & \textbf{13.7}$\pm$\textbf{0.2045} \\
Tooth root  & 9.4$\pm$1.5308 & 0.6$\pm$0.0516 \\
\bottomrule
\end{tabular}
\end{table}

\subsubsection{Component Ablation} To demonstrate the effectiveness of the proposed components of the ImplantFormer, i.e. convolutional stem and decoder, we conduct ablation experiments on both components to investigate their effects for ImplantFormer. When these components are removed, the patch embedding is generated on the input image and the multi-resolution feature maps are directly upsampled and element-wise added with the neighboring feature map. As discussed previously, we use $AP_{75}$ as the evaluation criteria in the following experiments.

The comparison results are shown in Table~\ref{table3}. We can observe from the table that the proposed components are beneficial for ImplantFormer, among which convolutional stem and decoder improves the performance by 0.9\% and 1.4\%, respectively. When combining both of these components, the improvement of performance reaches 2.9\%.

In Fig.~\ref{fig_vis_compare}, we visualize some examples of the detection results of different components for further comparison. We can observe from the figure that, for the ViT-base model, the predicted positions are relatively far away from the ground truth position. In the image of the first row, there is also a false positive detection at the right side. Both convolutional stem and decoder can reduce the distance between prediction and ground truth, thus improve the prediction accuracy. When combining convolutional stem and decoder, the model achieves more accurate predictions and reduces the number of missing or false positive detection.

\begin{figure}
\centering
\includegraphics[width=1.0\linewidth]{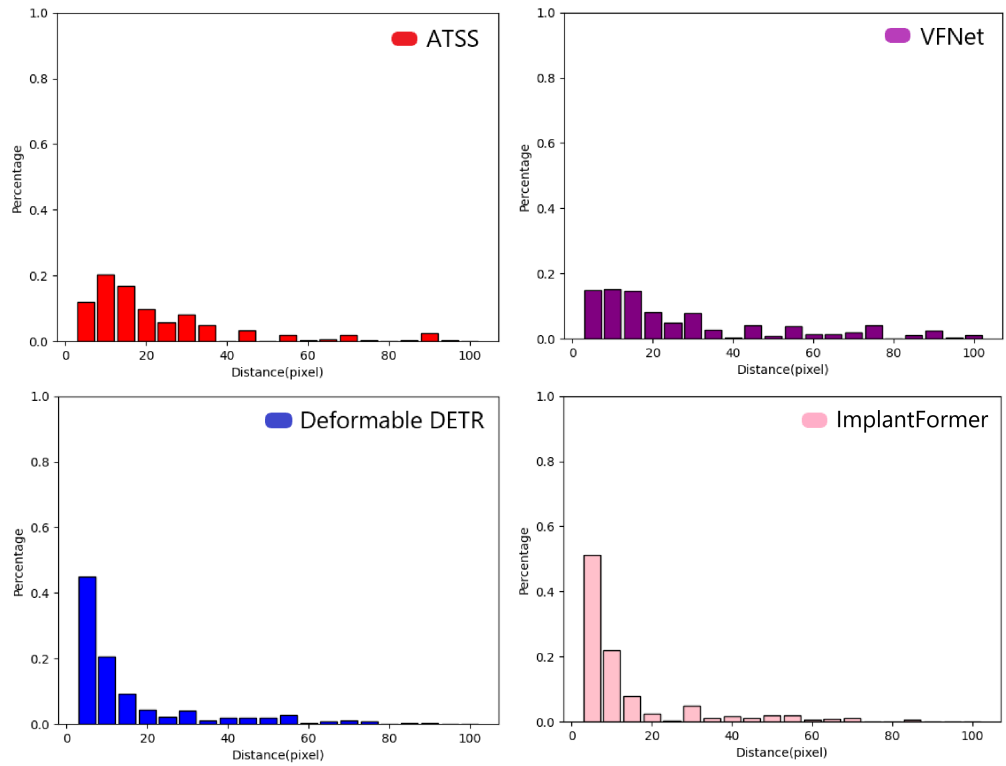}
\caption{The distribution of euclidean distances between the ground truth position and the predictions of different detectors.} \label{fig_distance}
\end{figure}

\begin{figure}
\centering
\includegraphics[width=1.0\linewidth]{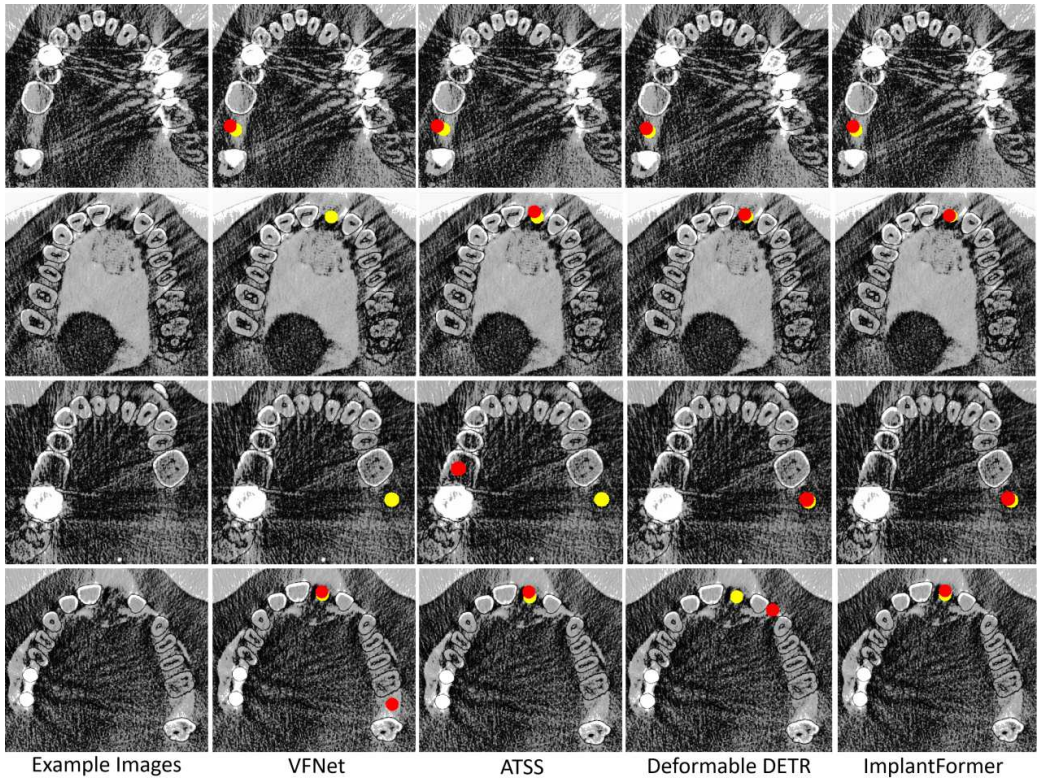}
\caption{Visual comparison of the detection results of different detectors. The red and yellow circles represent the predicted implant position and ground truth position, respectively.} \label{fig_vis_compare2}
\end{figure}

\begin{table}
\caption{The ablation study of convolutional stem and decoder.}\label{table3}
\centering
\begin{tabular}{cccc}
\toprule
Backbone                  & Conv stem & Conv decoder & $AP_{75}\%$          \\ \hline
\multirow{4}{*}{ViT-Base-ResNet-50} &           &              & 10.8$\pm$0.3491       \\
                          & $\checkmark$         &              & 11.7$\pm$0.4308 \\
                          &           & $\checkmark$            & 12.2$\pm$0.4065 \\
                          & $\checkmark$         & $\checkmark$      & \textbf{13.7}$\pm$\textbf{0.2045} \\
\bottomrule
\end{tabular}
\end{table}

\subsubsection{Comparison to The State-of-the-art Detectors} In Table~\ref{table4}, we compare the AP value of ImplantFormer with other state-of-the-art detectors. As no useful texture is available around the center of implant, where teeth are missing, the regression of the implant position is mainly based on the texture of neighboring teeth. Therefore, anchor-free methods (VFNet~\cite{zhang2021varifocalnet}, ATSS~\cite{zhang2020bridging}, RepPoints~\cite{yang2019reppoints}, CenterNet~\cite{zhou2019objects}) and transformer-based method (Deformable DETR~\cite{zhu2020deformable}) are more suitable for this regression problem. Nevertheless, we also employed two classical anchor-based detectors - Faster RCNN~\cite{ren2015faster} and Cascade RCNN~\cite{cai2018cascade} for comparison. Resnet-50 is employed as the feature extraction backbone for these detectors. To further determine the validity of vision transformer, we introduce ResNet-50 as the backbone for ImplantFormer, in which we remove the convolutional stem and the patch embedding operation, but keep the feature fusion block.

From Table~\ref{table4} we can observe that the anchor-based methods fail to predict the implant position, which confirms our concern. The transformer-based methods perform better than the CNN-based networks (e.g., Deformable DETR achieved 12.8\% AP, which is 0.7\% higher than the best-performed anchor-free network - ATSS). The ViT-based ImplantFormer achieves 13.7\% AP, which is 2.2\% and 0.9\% higher than ResNet-based ImplantFormer and Deformable DETR, respectively. The experimental results proved the effectiveness of our method, and the ViT-based ImplantFormer achieves the best performance among all benchmarks.

We choose two detectors from both anchor-free (e.g. ATSS and VFNet) and transformer-based (e.g. Deformable DETR and ImplantFormer) methods, respectively, to further demonstrate the superiority of ImplantFormer in the implant position prediction. Fig.~\ref{fig_distance} shows the euclidean distance between the ground truth position and the predictions of these detectors. The euclidean distances are summed in an interval of five. Smaller the distance, more accurate the implant position prediction. From the figure we can observe that, for anchor-free detection methods, the distance distributes equally from 0 to 30 pixels. Only 35\% predictions distribute within 10 pixels from groundtruth position. In contrast, for transformer-based detection methods, the distance mainly distributes in the range of 0 to 20 pixels. More than 70\% of the predictions located within 10 pixels from groundtruth. Considering that the diameter of implant is around 20 pixels, the predictions with distance more than 10 pixels are meaningless. Therefore, the transformer-based detectors are more suitable for implant prediction task. Compared to Deformable DETR, the ImplantFormer generates about 8\% more predictions with distance less than 5 pixels, which indicates that the ImplantFormer achieve much better performance even its AP is only 1.2\% higher.

In Fig.~\ref{fig_vis_compare2}, we also visualize the detection results of these detectors for four example images. We can observe from the first row of the figure that the transformer-based methods perform better than the anchor-free methods, which is consistent with the distance distribution. As shown in the second and third row, the anchor-free methods generate false detections, while the transformer-based methods perform accurately. In the last row, an example of hard case is given, where the patients teeth are sparse in several places, which leads to error detection of VFNet and Deformable DETR, while the proposed ImplantFormer still performs accurately.

As shown in the second row, an example of hard case is given, where the patient's teeth are sparse in several places, which leads to error detection of VFNet and Deformable DETR, while the proposed ImplantFormer still performs accurately.

\begin{figure}
\centering
\includegraphics[width=1.0\linewidth]{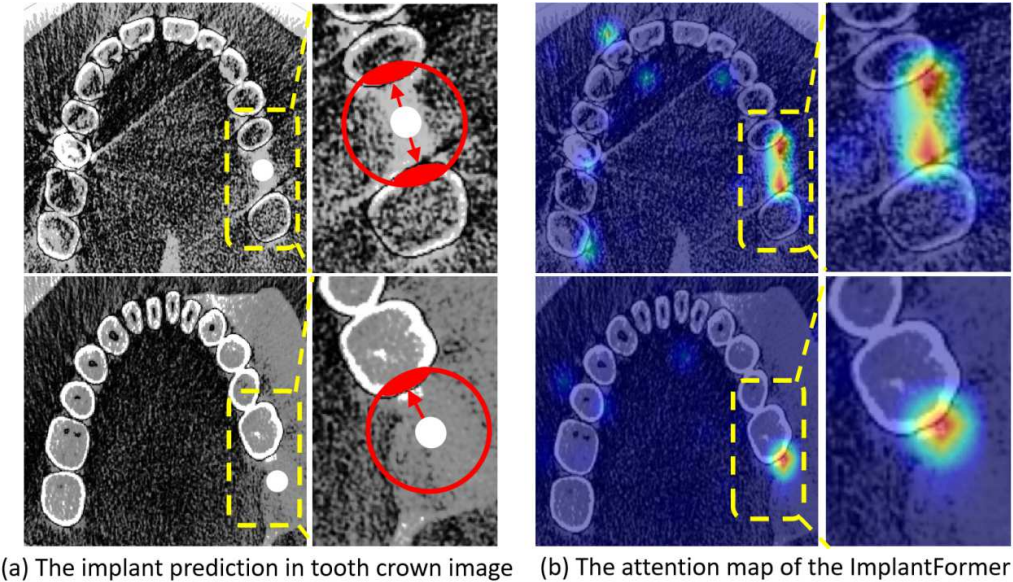}
\caption{Visual comparison of the location mechanism of implant position. (a) The groundtruth position in tooth crown image. The red areas in the enlarged rectangle represent the edge of neighboring teeth. (b) The attention map of the ImplantFormer, where the highlighted rectangle shows the attention region of the network.} \label{fig_attention}
\end{figure}

\begin{table}
\caption{Verification of the proposed method and other state-of-the-art detectors on our dataset. "-" represents 0.}\label{table4}
\centering
\begin{tabular}{cccc}
\toprule
Methods                                                & Network         & Backbone                   & $AP_{75}\%$    \\ \hline
\multicolumn{1}{c}{\multirow{2}{*}{Anchor-based}}      & Faster RCNN     & \multirow{8}{*}{ResNet-50} & -    \\
\multicolumn{1}{c}{}                                   & Cascade RCNN    &                            & -    \\ \cline{1-2}
\multicolumn{1}{c}{\multirow{5}{*}{Anchor-free}}       & CenterNet       &                            & 10.9$\pm$0.2457  \\
\multicolumn{1}{c}{}                                   & ATSS            &                            & 12.1$\pm$0.2694 \\
\multicolumn{1}{c}{}                                   & VFNet           &                            & 11.8$\pm$0.8734 \\
\multicolumn{1}{c}{}                                   & RepPoints       &                            & 11.2$\pm$0.1858  \\
\multicolumn{1}{c}{}                                   & ImplantFormer          &                            & 11.5$\pm$0.3748 \\ \cline{1-2}
\multicolumn{1}{c}{\multirow{2}{*}{Transformer-based}} & Deformable DETR &                            & 12.8$\pm$0.1417 \\ \cline{3-3}
\multicolumn{1}{c}{}                                   & ImplantFormer          & ViT-Base-ResNet-50                   & \textbf{13.7}$\pm$\textbf{0.2045} \\
\bottomrule
\end{tabular}
\end{table}

\begin{figure}
\centering
\includegraphics[width=0.8\linewidth]{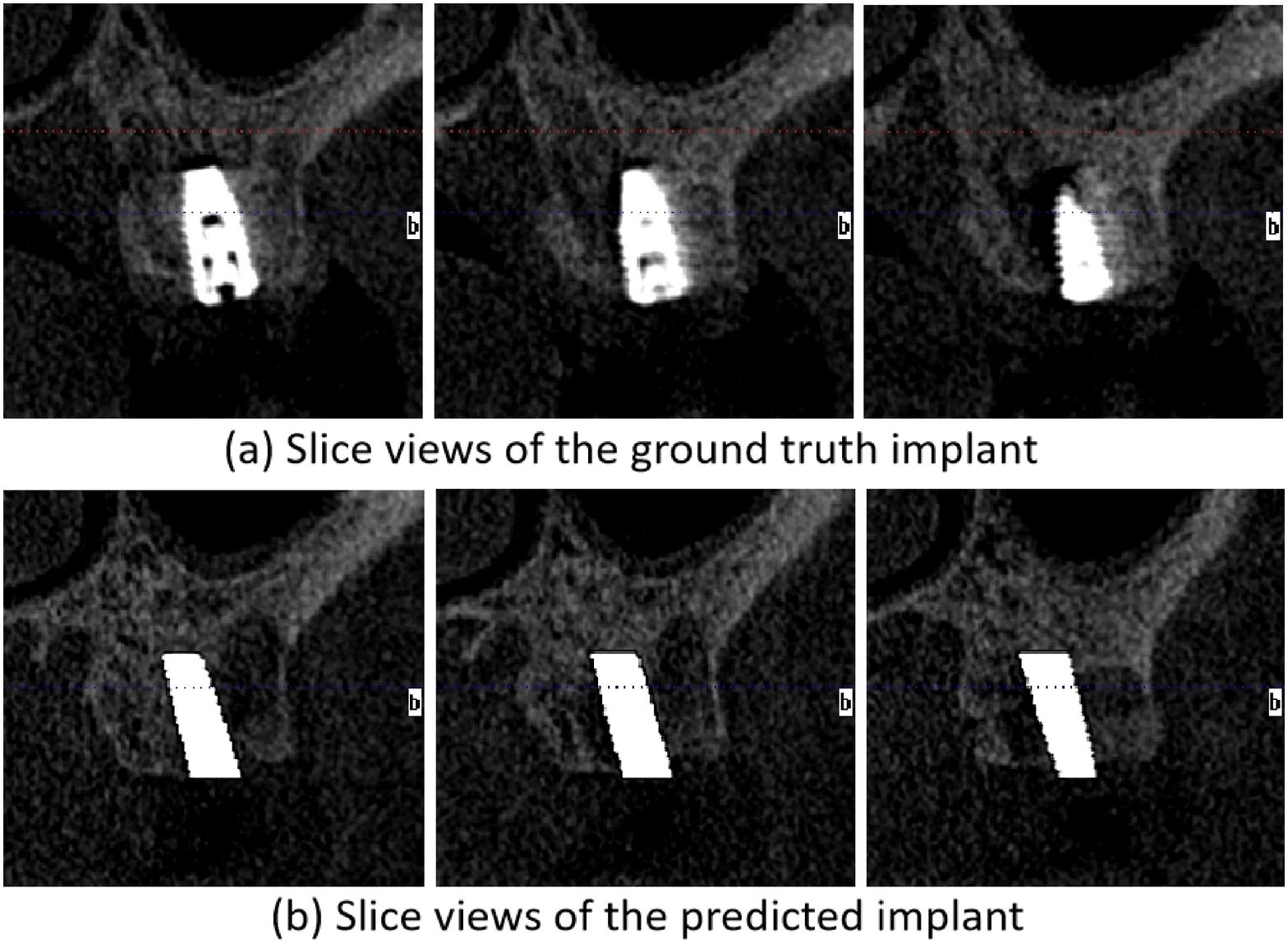}
\caption{Visual comparison of the ground truth implant (first row) and the implant predicted by ImplantFormer (second row) in different slice views.} \label{fig_patient_implant}
\end{figure}

\subsubsection{Visualization of Attention Map} To verify whether the location mechanism of implant position of the ImplantFormer is in line with the dentist, we visualize the attention map of the ImplantFormer in the Fig.~\ref{fig_attention}. The enlarged portion in Fig.~\ref{fig_attention}(a) shows the location mechanism of implant position by the dentist, which is determined by the edge of neighboring teeth. Fig.~\ref{fig_attention}(b) is the attention map of ImplantFormer, from which we can observe that the network attention is on the edges of neighboring teeth of implant position, which is in accordance with the location mechanism of implant by the dentist. Simultaneously, the network attention around the neighboring teeth also indicates that the vision transformer can establish the relationships between long-ranged pixels, i.e. from the implant position to the neighboring teeth.

\subsubsection{Comparison of Slice Views of Implant} To validate the effectiveness of the proposed ImplantFormer, we compare four slice views of the actual implant with the predicted implant in Fig.~\ref{fig_patient_implant}. The slice view is the different longitudinal views of CBCT data of the implant, which can well demonstrate the direction and placement of the implant. For the predicted implant, a cylinder with a radius of 10 pixels centered at $Pos_r$, is generated, and the implant depth is manually set. The pixel value of CBCT data inside the cylinder area is set as 3100. For fair comparison, the longitudinal views from the CBCT data of both ground truth and predicted implant are selected at the same position.

From the figure, we can observe that the implant position and direction generated by ImplantFormer are consistent with the ground truth implant, which confirms the accuracy of the proposed ImplantFormer.

\section{Conclusions}
In this paper, we introduce a transformer-based Implant Position Regression Network (ImplantFormer) for CBCT data based implant position prediction, which includes both local context and global information for more robust prediction. We creatively propose to train ImplantFormer using tooth crown images by projecting the annotations from tooth root to tooth crown via the space transform algorithm. In the inference stage, the outputs of ImplantFormer will be projected back to the tooth root area as the predicted positions of the implant. Visualization of attention map and slice view of implant demonstrated the effectiveness of our method and qualitative experiments show that the proposed ImplantFormer achieves superior performance than the state-of-the-art object detectors.



\bibliographystyle{splncs04}
\bibliography{ref}

\end{document}